\documentclass[pdflatex,sn-basic,iicol]{sn-jnl}

\usepackage{amsmath,amssymb,amsfonts}
\usepackage{graphicx}
\usepackage{textcomp}
\usepackage{xcolor}
\usepackage{booktabs}
\usepackage{bbm}
\usepackage{subfigure}
\def\BibTeX{{\rm B\kern-.05em{\sc i\kern-.025em b}\kern-.08em
		T\kern-.1667em\lower.7ex\hbox{E}\kern-.125emX}}
\usepackage{multirow}
\usepackage{soul}
\usepackage{color}

\jyear{2021}%

\theoremstyle{thmstyleone}%
\theoremstyle{thmstyletwo}%

\theoremstyle{thmstylethree}%

\begin{document}

\title[\\]{Decoupled Knowledge with Ensemble Learning for Online Distillation}


\author[1]{\fnm{Baitan} \sur{Shao}}\email{shaoeric@foxmail.com}

\author*[1]{\fnm{Ying} \sur{Chen}}\email{chenying@jiangnan.edu.cn}
\equalcont{These authors contributed equally to this work.}

\affil*[1]{\orgdiv{Key Laboratory of Advanced Process Control for Light Industry Ministry of Education}, \orgname{Jiangnan University}, \orgaddress{\street{Lihu Avenue}, \city{Wuxi}, \postcode{214122}, \state{Jiangsu}, \country{China}}}


\abstract{Offline knowledge distillation is a two-stage pipeline that requires expensive resources to train a teacher network and then distill the knowledge to a student network for deployment. Online knowledge distillation, on the other hand, is a one-stage strategy that alleviates the requirement with mutual learning and collaborative learning. Recent peer collaborative learning (PCL) integrates online ensembling, collaboration of base networks (student) and temporal mean teacher (teacher) to construct effective knowledge. However, the model collapses occasionally in PCL due to high homogenization between the student and the teacher. In this paper, the cause of the high homogenization is analyzed and the solution is presented. A decoupled knowledge for online knowledge distillation is generated by an independent teacher, separate from the student. Such design can increase the diversity between the networks and reduce the possibility of model collapse. To obtain early decoupled knowledge, an initialization scheme for the teacher is devised, and a 2D geometry-based analysis experiment is conducted under ideal conditions to showcase the effectiveness of this scheme. Moreover, to improve the teacher's supervisory resilience, a decaying ensemble scheme is devised. It assembles the  knowledge of the teacher to which a dynamic weight which is large at the begining of the training and gradually decreases with the training process, is assigned. The assembled knowledge serves as a strong teacher during the early training and the decreased-weight-assembled knowledge can eliminate the distribution deviation under the potentially overfitted teacher's supervision. A Monte Carlo-based simulation is conducted to evaluate the convergence. Extensive experiments on CIFAR-10, CIFAR-100 and TinyImageNet show the superiority of our method. Ablation studies and further analysis demonstrate the effectiveness. The code is available at \url{https://github.com/shaoeric/Decoupled-Knowledge-with-Ensemble-Learning-for-Online-Distillation}.}




\maketitle

\section{Introduction}\label{sec1}

Over the past few years, deep learning has gradually come to dominate the field of computer vision. Mainstream tasks in computer vision, like image classification \cite{ning2022hcfnn} and object detection \cite{zaidi2022survey}, have also been surprisingly successful with the help of deep learning techniques. Cumbersome networks tend to get better feature extraction performance in these tasks because of their powerful representation capabilities. However, considering the real-time performance of the system and the user experience, such a cumbersome network can hardly meet the product requirements of industry. To solve this problem, model compression techniques have received a lot of attention from researchers. The popular model compression techniques include model pruning \cite{jiang2022model}, model quantization \cite{becking2022ecq}, knowledge distillation \cite{beyer2022knowledge} and lightweight module design \cite{sivapalan2022annet}.

Knowledge distillation (KD) provides an effective learning method to improve a compact student network by mimicking the outputs from a cumbersom teacher with better performance. Hinton et al. \cite{hinton2015KD} introduces distillation temperature to soften the class probability distributions from the teacher network for more effective supervision on the student. Based on this idea, a variety of knowledge from the teacher have been developed to optimize the student, such as hint representations \cite{Romero2015FitNetsHF}, attention-based feature maps \cite{komodakis2017paying}, relational information \cite{park2019relational}, contrastive representation \cite{tian2019crd} and so on. The above knowledge and methods follow a two-stage training pipeline and demand a teacher network with better performance, usually named as traditional offline knowledge distillation. However, for a resource-constrained training task, there may be neither an existing teacher network available nor sufficient time and devices to train a teacher network.

\begin{figure*}[htbp]
	\centering
	\includegraphics[width=\linewidth]{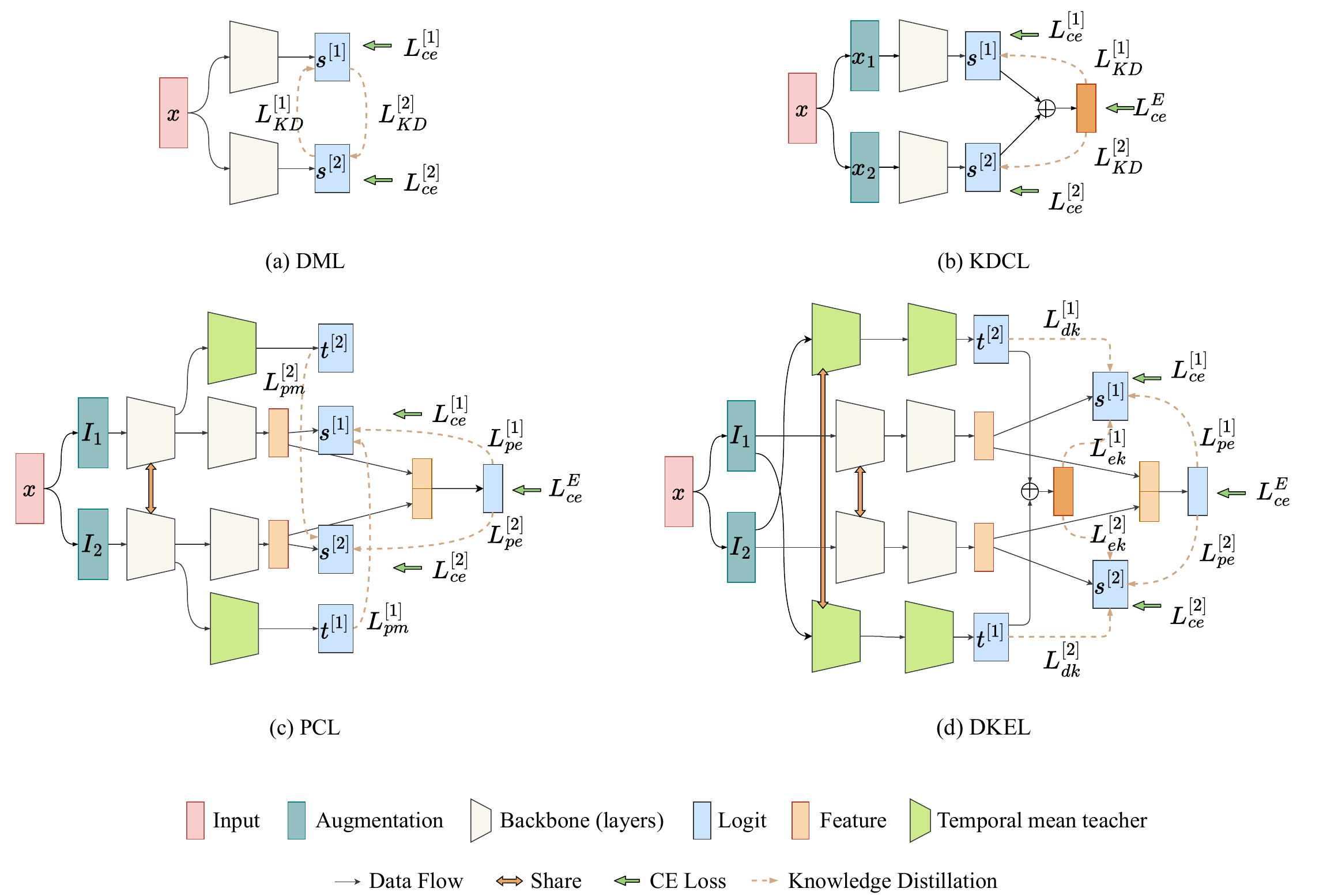}
	\caption{Illustration of four online distillation methods. (a) Deep mutual learning (DML). (b) Online distillation via collaborative learning (KDCL). (c) Peer collaborative learning (PCL). (d) Decoupled knowledge with ensemble learning (DKEL).}
	\label{fig:overview}	
\end{figure*}

To alleviate the problem, online knowledge distillation is developed to optimize multiple compact networks simultaneously in a one-stage end-to-end training procedure. Deep mutual learning (DML) \cite{zhang2018deep} (Fig. \ref{fig:overview}(a)) demonstrates that it is feasible for models to learn from each other instead of the blind leading the blind, and models in such an online manner obtain better performance than those learning alone. This conclusion has inspired much research in collaborative learning. Ensemble learning, widely used in existing online KD, such as KDCL \cite{guo2020online}  (Fig. \ref{fig:overview}(b)) , builds a robust pseudo teacher network by reducing the variance of each individual network, which in turn alleviates the problem of inaccurate supervision in the early period of online KD. Some recent work trends to name an individual branch of a network as a peer \cite{chen2020online, wu2021peer}. For constructing more robust and stable supervision, temporal mean teacher (teacher) is introduced by peer collaborative learning (PCL) to improve the base network (student) \cite{wu2021peer} (Fig. \ref{fig:overview}(c)), essentially a technique that applies an exponential moving average (EMA) \cite{nakano2017generalized} over the network parameters, which is applied in batch normalization \cite{garbin2020dropout} and YOLO \cite{wang2022yolov7} as well.

In this work, with PCL as the baseline approach, we propose decoupled knowledge with ensemble learning (DKEL), as illustrated in Fig. \ref{fig:overview}(d). Our proposed method, DKEL, incorporates two forms of transferred knowledge, namely decoupled knowledge and decaying ensemble knowledge. Decoupled knowledge is introduced to address the issue of model collapse in PCL, with a detailed analysis conducted to identify the underlying cause. We propose a solution by expanding the solution space of the teacher, which involves constructing an independent teacher and designing a teacher initialization scheme to generate early decoupled knowledge. An ideal experiment is conducted to validate this approach, with the results presented in a 2D geometrical format. Furthermore, to facilitate strong and effective supervision, we devise a decaying ensemble scheme where the logits of teacher peers are assembled and assigned a decaying weight for student supervision. Both the decoupled and decaying ensemble knowledge generated by the teacher enhance the effectiveness of supervised information for the student, while the teacher's performance benefits from EMA updates.

Our contributions are summarized as follows:
\begin{itemize}
	\item[$\bullet$] Our method for online distillation proposes decoupled knowledge with ensemble learning, and we show its theoretical and empirical superiority over existing methods.
	\item[$\bullet$] A teacher network with decoupled knowledge is constructed to deal with model collapse caused by high homogenization of the teachers and the student's peers.
	\item[$\bullet$] A scheme of decaying ensemble knowledge is designed, which serves as an early robust teacher to accelerate the optimization and avoids providing overfitting supervision during the late training.
	\item[$\bullet$] The proposed method is evaluated on multiple datasets over different architectures. The performance is reported compared with the SOTAs. 
\end{itemize}

\section{Related work}\label{sec2}

Knowledge distillation is usually applied to model compression, wich is divided  into offline and online knowledge distillation \cite{gou2021knowledge}. 

\textbf{Offline knowledge distillation} demands a pretrained teacher network and a student network, and the student learns from the teacher and the ground truth simultaneously. Hinton et al. put forward the concept of knowledge distillation and proposed to transfer knowledge from a cumbersome teacher to a compact student by aligning the soft distribution of the teacher and the student \cite{hinton2015KD}. In general, offline knowledge distillation tends to train a student network with surprisingly good performance and is very effective for compressing a large network, however, it usually follows a two-stage paradigm, which greatly increases the training time and computational overhead. 

\textbf{Online knowledge distillation} is a one-stage end-to-end training method, which transfers knowledge among multiple networks in a mutual manner with no need for extra time and computing resources to pretrain a cumbersome teacher network. Zhang et al. pioneer a deep mutual learning method to explore the feasibility of online knowledge distillation \cite{zhang2018deep}, which distills knowledge among multiple parallel models with the same input. For maintaining the diverisity of the multiple networks, Guo et al. randomly augment the same input for each individual network and aggregate all the output logits into an ensemble soft label for optimizing each network in the online distillation \cite{guo2020online}. Wu et al. aggregate with stacking strategy and utilize temporal mean teacher to derive a robust prediction for training and inference \cite{wu2021peer}.

\section{Methodology}
\subsection{Formulation}
Given a $C$-category-object dataset of $N$ training samples $D=\left\{ (x_i,y_i) \right\}_i^N$, where $x_i$ is the $i$-th image and $y_i$ is the corresponding ground truth, and $y_i\in \left\{ 1,2,\cdots,C\right\}$. Feeding the samples into the teacher and the student, the corresponding derived output logits are denoted as $t_i=[t_i^1,t_i^2,\cdots, t_i^C]$ and $s_i=[s_i^1,s_i^2,\cdots, s_i^C]$. The softmax function $\sigma(\cdot)$ normalizes logits into a probability vector, and Kullback-Leibler (KL) divergence is usually used to minimize the probability distribution gap between the teacher and the student. In general, the probability distribution needs to be smoothed by a distillation temperature hyperparameter $\tau$. Specifically, the knowledge distillation loss function is denoted as:
\begin{equation}
	L_{KD}(s, t) = \frac{1}{N}\tau^2 \sum_{i=1}^N KL(\sigma(\frac{t_i}{\tau}), \sigma(\frac{s_i}{\tau})   ).
\end{equation}
The value of the KL function will be 0 if and only if $t_i=s_i$.

\subsection{PCL}

This section provides a brief description of the PCL method. It integrates two types of models into a unified framework: a base model and a temporal mean teacher, in which the base model is regarded as a student $s$ and the temporal mean teacher is abbreviated as teacher $t$.

Let $s^{[p]}$ denote the $p$-th peer of the student and $t^{[p]}$ denotes the corresponding teacher peer. As illustrated in Fig. \ref{fig:overview}(c), PCL augments the input $x$ to $I_p$ for the $p$-th peer, and derives the corresponding flatten feature and logits. The features are assembled by stacking as the peer ensemble $E$, which supervises each logits with $L_{pe}$, like $L_{KD}$ does. The ground truth optimizes logits with cross entropy loss $L_{ce}$ to avoid the blind guiding the blind and optimizes the ensemble $E$ with $L_{ce}^E$ to improve the capacity. Besides, the teacher peers with better generalisation are employed to transfer knowledge via $L_{pm}$. 

Assuming that $m$ denotes the number of peers, the loss function of PCL for $s$ can be expressed as:
\begin{equation}
	L_{PCL} = \sum_{p=1}^{m}(L_{ce}^{[p]}+L_{pe}^{[p]} + L_{pm}^{[p]}) + L_{ce}^{E},
\end{equation}
where
\begin{align}
	L_{pe}^{\left[ p \right]}&=L_{KD}\left( s^{\left[ p \right]},E \right) \nonumber \\
	&=L_{KD}\left( s^{\left[ p \right]},\rho \left( \left\{ s^{\left[ j \right]} \mid j=1\cdots m \right\} \right) \right), \label{pe}
\end{align}
where $\rho(\cdot)$ is an ensemble function, and
\begin{equation}
	L_{pm}^{[p]} = \frac{1}{m-1}\sum_{j=1,j\neq p}^{m}L_{KD}^{[j]}(s^{[p]}, t^{[j]}). \label{pm}
\end{equation}

Following each iteration of the student, the teacher's parameters are updated using EMA:
\begin{equation}
	t^{\prime} = \eta s^{\prime} + (1-\eta) t,
\end{equation}
where $\eta \in (0,1)$ is a hyperparameter, $s^{\prime}$ and $t^{\prime}$ are the optimized student and teacher, respectively.

\subsection{Decoupled knowledge}
\begin{figure}[htbp]
	\centering
	\includegraphics[width=0.9\linewidth]{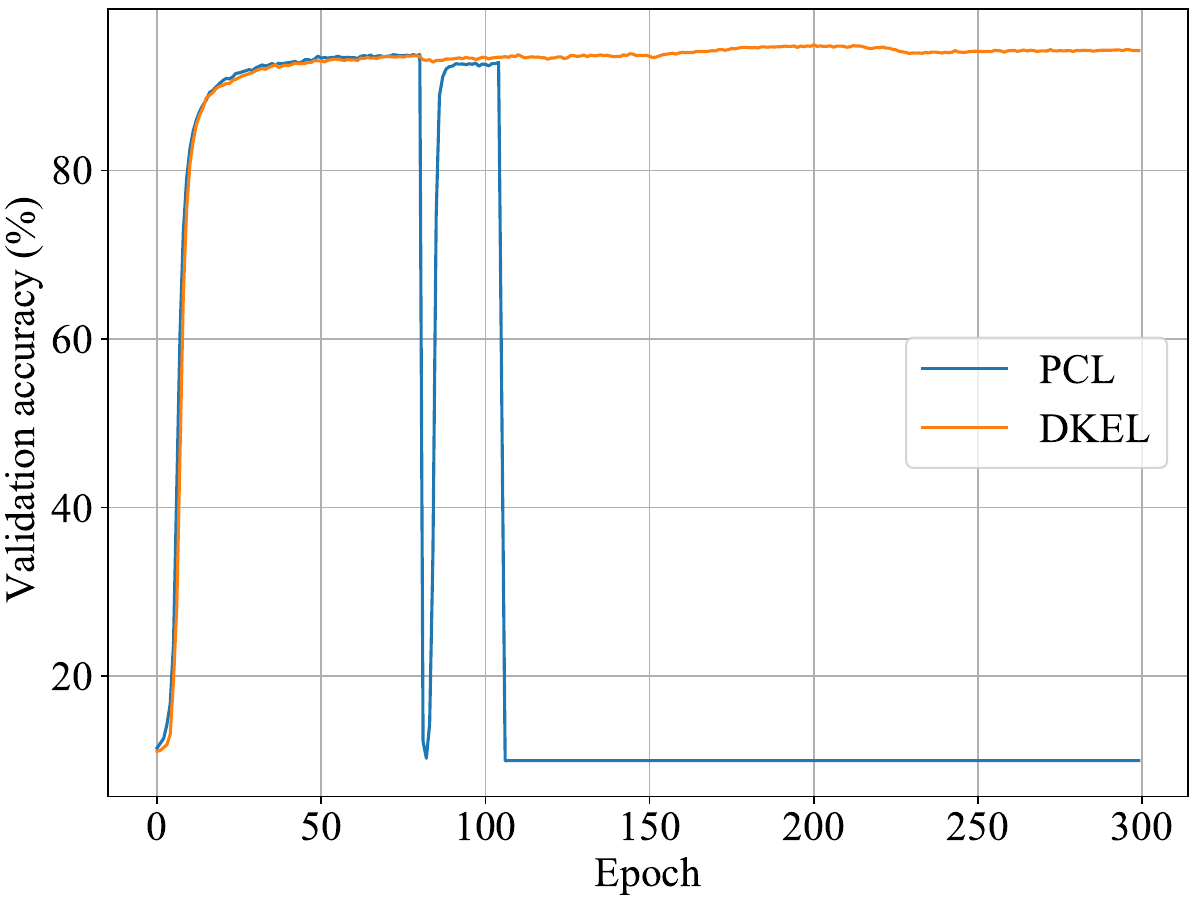}
	\caption{Validation accuracy curves of VGG-16 on CIFAR-10.}
	\label{fig:collapse}	
\end{figure}

Decoupled knowledge focuses on the problem of potential model collapse in PCL \cite{wu2021peer}. As illustrated in Fig. \ref{fig:collapse}, PCL triggers certain situations which may cause the model training collapse. The parameter matrices of the collapsed networks are checked and it is found that in order to keep the distribution of teachers' and students' knowledge as close as possible, all parameters in the collapsed network are forced to be zero and the logits are all zero vectors. Formularly, $\forall x\in \{x_i\}_{i=1}^N$, s.t. $s_x^{[p]}=t_x^{[p]}=0$, where network $s$ and $t$ both act as linear transformation with full zero parameters.

The essential reason for collapse is that the homogenization of the teacher's and the student's peers (abbreviated as $s\&t$ homogenization) is too high, resulting in too small distillation loss. Furthermore, in the $s\&t$ homogenization situation, the logits values are small due to positive feedback, resulting in small cross-entropy loss. To summerize, the $s\&t$ homogenization will cause too small $L_{PCL}$. At this time, the L2 regularization term in the SGD optimizer is prioritized for optimization, which causes smaller and smaller absolute values of all the parameters until the network collapses.

The $s\&t$ homogenization mainly results from the fact that they share the same backbone, while independent peers, due to their relatively small solution space, can easily trigger the conditions for collapse mentioned above.

\begin{figure}[htbp]
	\centering
	\includegraphics[width=\linewidth]{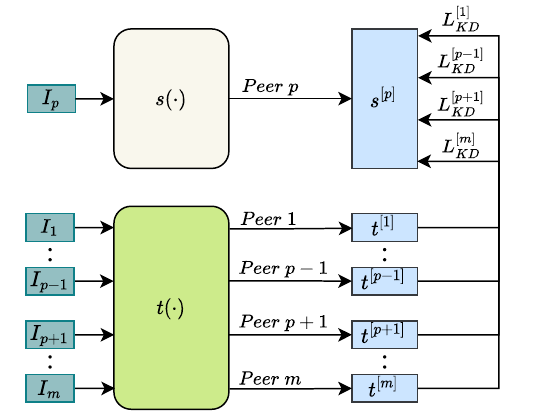}
	\caption{Illustration of decoupled knowledge transferred to the student's $p$-th peer.}
	\label{fig:decoupled}
\end{figure}

In order to reduce the homogenization, a simple but effective decoupled $s\&t$ strategy is designed in which an independent teacher is constructed to transfer decoupled knowledge to the student and the teacher shares no parameters with the student. As shown in Fig. \ref{fig:decoupled}, the augmentation $I_p$ is fed into the student backbone $s\left(\cdot\right)$ and the $p$-th peer sequentially, the logits $s^{\left[p\right]}$ is derived. For the teacher, all augmentations except the $p$-th are fed to the teacher backbone $t\left(\cdot\right)$ and its $m-1$ peers to get the corresponding logits set $\left\{ t^{\left[ j \right]} \mid j=1\cdots m,j\ne p \right\} $, therefore the loss function of the decoupled knowledge for $s^{\left[p\right]}$ is denoted as:

\begin{equation}
	L_{dk}^{[p]} = \frac{1}{m-1}\sum_{j=1,j\neq p}^{m}L_{KD}^{[j]}(s^{[p]}, t^{[j]}). \label{dk}
\end{equation}

\begin{figure*}[htbp]
	\centering
	\includegraphics[width=\linewidth]{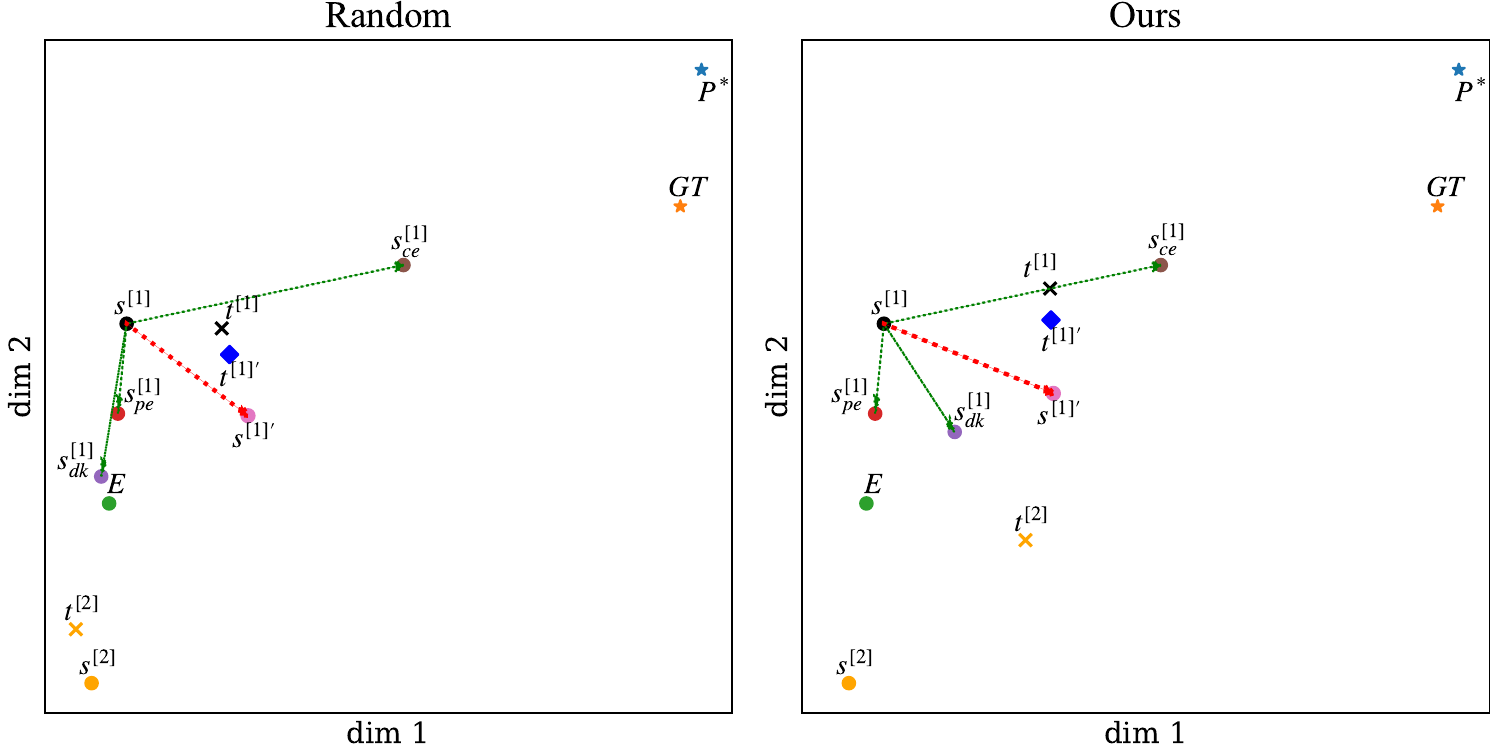}
	\caption{Distribution illustration comparision between two initiations.}
	\label{fig:tool}	
\end{figure*}

Following the construction of the decoupled knowledge, it is necessary to consider how to initialize the teacher network. Random initialization is widely used in current neural networks, however, it can cause early optimization direction bias, which is called the early knowledge bias in this work. To make it clear, as shown in Fig. \ref{fig:tool}, we design an ideal experiment and demonstrate the result in 2D geometrical style. The distributions of data and network are abstractly represented as points and the distance between two points is idealized to represent their distribution gap. $P^{*}$ and $GT$ represent the real data distribution and ground truth distribution, respectively. $s^{[p]}$ represents the $p$-th peer's distributions of the student, and $t^{[j]}$ is for the teacher's $j$-th peer distribution. $E$ is the stack ensemble of the two student peers, which is at the midpoint of $s^{[1]}$ and $s^{[2]}$. For clarity of the figure representation, only the optimizations of $s^{[1]}$ and $t^{[1]}$ are shown. $s^{[1]}$ is optimized under three supervisions: $GT$ optimizes $s^{[1]}$ to $s_{ce}^{[1]}$ with $L_{ce}^{[1]}$, $E$ optimizes $s^{[1]}$ to $s_{pe}^{[1]}$ with 
$L_{pe}^{[1]}$ and  $t^{[2]}$ optimizes $s^{[1]}$ to $s_{dk}^{[1]}$ with $L_{dk}^{[1]}$.

According to the law of vector summation, the optimized peer distribution $s^{[1]^{\prime}}$ is denoted as:
\begin{equation}
	s^{[1]^{\prime}} = s^{[1]} + \overrightarrow{s^{[1]}s_{ce}^{[1]}} + \overrightarrow{s^{[1]}s_{pe}^{[1]}} + \overrightarrow{s^{[1]}s_{dk}^{[1]}}.
\end{equation}
Then, EMA optimizes $t^{[1]}$ to $t^{[1]^{\prime}}$, which is denoted as:
\begin{equation}
	\label{ema}
	t^{[1]^{\prime}} = \eta s^{[1]^{\prime}} + (1-\eta) t^{[1]}.
\end{equation}

Due to the randomness of the initialization, it is possible that, $t^{[2]}$ and $P^{*}$ are on the opposite sides of $s^{[1]}$, i.e., $\angle t^{[2]}s^{[1]}P^{*}$ is an obvious obtuse angle as well as $\angle s_{dk}^{[1]}s^{[1]}P^{*}$. It will cause the optimized $s^{[1]^{\prime}}$ to bias away from $P^{*}$ so that the optimized $t^{[1]^{\prime}}$ will suffer from the bias. 

Therefore, a scheme of initialization for the teacher is proposed, which includes two steps: 

1) copy the student's weights to the teacher in order to ensure the same initial distribution of the two networks,

2) optimize the teacher with cross entropy in only few steps with small learning rate.

Such an initialization tunes $t^{[2]}$ to the position on the same side of $P^{*}$ with $s^{[1]}$ as reference, as shown in Fig. \ref{fig:tool}(b), which accelerates $s_{dk}^{[1]}$ to approach $P^{*}$ in direction, compared with the $s_{dk}^{[1]}$ using random initialization. Benefiting from this, the optimized $s^{[1]^{\prime}}$ and $t^{[1]^{\prime}}$ are closer than that of the random initialization, which improves performance.

Additionally, from a statistical point of view, $t$ using the EMA method is equivalent to a first-order moment estimate of the $s$ parameter as well as momentum of Adam optimizer \cite{xie2022adaptive}, and if the initial performance of $t$ can be improved, it will also improve the performance of $t$ globally during the training.

\subsection{Decaying ensemble strategy}

Ensemble learning can be used to enhance the generalization performance of multiple networks and obtain better evalution. It is commonly applied to construct a robust supervised information with multiple logits in the field of online knowledge distillation. The teachers in PCL and the proposed decoupled knowledge directly supervise the students' training without the logits ensemble, which can lead to students' performance limitation due to the incapacity of the early teachers. 
As the networks are continuously optimized, the teacher's peers will gradually fit or even overfit to $GT$. Although its ensemble usually yields the best-generalization knowledge, there is still a deviation towards the $P^{*}$ distribution, which may provide overfitting supervision for the student.

Based on the above statement, a decaying ensemble strategy for the teacher is further proposed to improve the decoupled knowledge. For the $p$-th student peer to be distilled, the loss function of $s^{[p]}$ and the ensemble of the teacher's peers can be denoted as:

\begin{equation}
	\label{ek}
	L_{ek}^{[p]}=L_{KD}(s^{[p]}, \bar{t}^{[p]}).
\end{equation}
where
\begin{equation}
	\label{t_ensemble}
	\bar{t}^{[p]} = \frac{1}{m-1}\sum_{j=1,j\neq p}^m t^{[j]},
\end{equation}
where the purpose of $j\ne p$ is to avoid high homogenization and potential model collapse.

In order to reduce distribution deviation of the overfitted teacher towards $P^{*}$, an exponential weight function is applied to $L_{ek}^{[p]}$ to decay with the increasing epoch $e$, which is denoted as:
\begin{equation}
	\omega(e)=exp(-\gamma \cdot e)
\end{equation}
where $\gamma$ is a positive decaying hyperparameter. Therefore loss function of the proposed DKEL for $s$ is denoted as:
\begin{equation}
	\label{loss}
	L = \sum_{p=1}^{m}[L_{ce}^{[p]}+L_{pe}^{[p]} +\omega(e)L_{ek}^{[p]} + (1-\omega(e))L_{dk}^{[p]}] + L_{ce}^{E}.
\end{equation}
\begin{figure}[h]
	\centering
	\includegraphics[width=\linewidth]{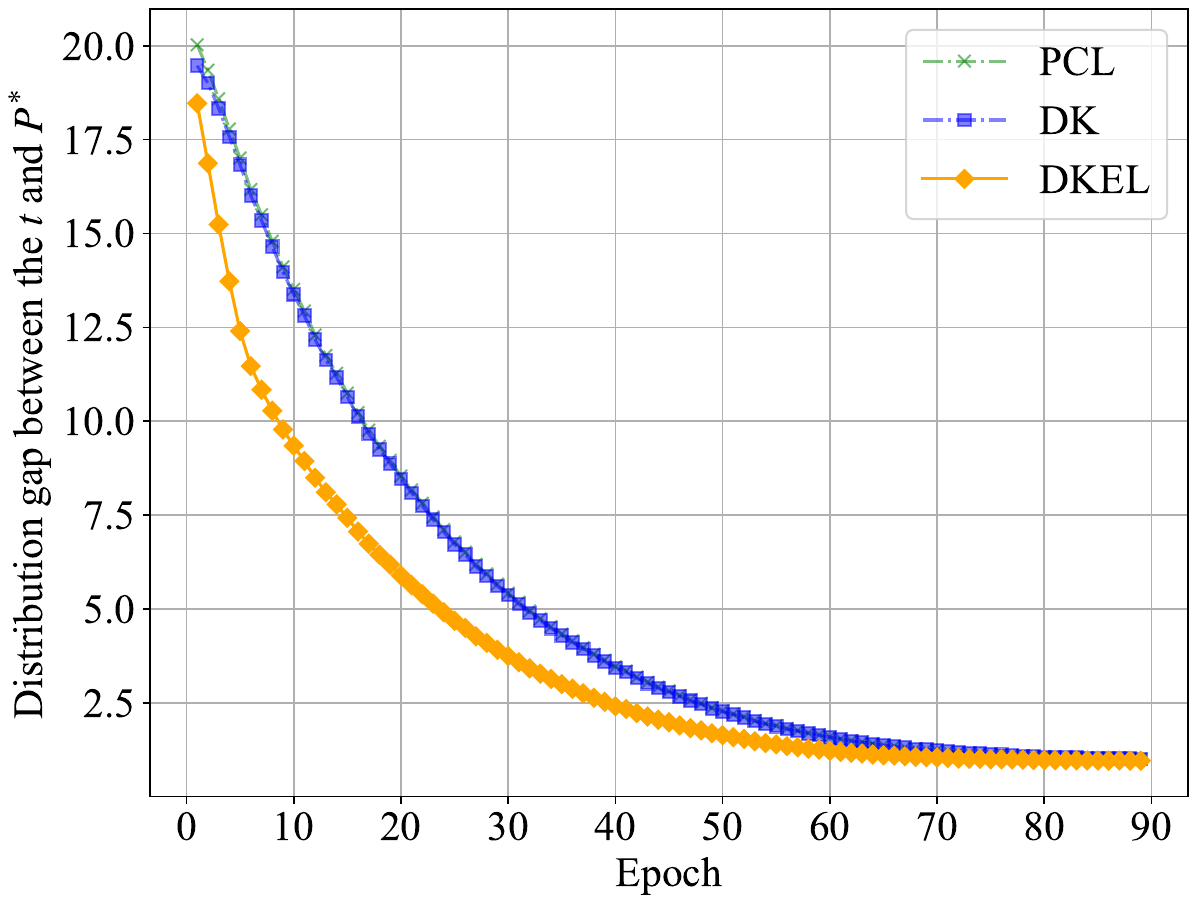}
	\caption{Illustration of distribution gaps with the Monte Carlo simulation. "DK" denotes the proposed decoupled knowledge. Only one peer's results are reported.}
	\label{fig:trials}	
\end{figure}

A Monte Carlo simulation using three network peers are designed to verify the aboved idea based on the ideal experiment mentioned in the previous section. Supposing that distributions are represented as points, the distance between two points represents the loss of their corresponding distributions, the optimization direction is a vector direction from the point to the target point, and the ensemble is expressed as the midpoint of two points. Here, the learning rate is 0.1, only 1 step optimization is for constructing the decoupled knowledge, and $\eta=0.5$ for EMA. 90 epochs of experment is conducted, each containing 10,000 trials, in order to compare the distribution gaps between $t$ and $P^{*}$, and we report the average gap for each epoch. As shown in Fig. \ref{fig:trials}, the blue dot line indicates the proposed decoupled knowledge, abbreviated as DK in the figure, the green dot line and the orange line indicate PCL and DKEL, respectively. The results show that DK only slightly improves upon PCL, and both methods converge at around the 70th epoch. However, DKEL utilizes decoupled knowledge to speed up the optimization process during the first 10 epochs by creating strong supervision. This allows the network to converge faster, around the 60th epoch. Additionally, the decaying ensemble scheme in DKEL produces a smaller gap compared to the other two methods.

Based on the above descriptions, Algorithm \ref{algorithm} describes the optimization process. After optimization, inference and deployment are performed using the teacher without extra overload because of the identical model structure of the teacher and the student.
\begin{algorithm}
\begin{algorithmic}[1]
	\caption{Decoupled Knowledge with Ensemble Learning for Online Distillation.}
	\label{algorithm}
	\Require Training data ${(x_i,y_i)}_{i=1}^N$; A $m$-peer student to be trained $s$.
	\Ensure A $m$-peer teacher network $t$ with the same architecture as $s$.
	\State Randomly initialize the student parameters.
	\State Make a deep copy of the student as the decoupled teacher, ensuring their parameters are the same.
	\State Optimize all peers of $t$ with cross entropy loss in only few steps with small learning rate to construct decoupled knowledge.
	\For{$e=0\rightarrow Epoch_{max}$}
		\State Randomly transform $x_i$ to get augmentations $\{x_{i}\}_{p=1}^m$ for each peer.
		\State Feedforward and get features and logits of the $s$ peers.
		\State Assemble features as the peer ensemble feature $E$. 
		\State Feedforward and get logits of the $t$ peers. 
		\State Compute loss $L_{ce}^{[p]}$ of the $s$ peers. 
		\State Compute loss $L_{ce}^E$ of feature $E$. 
		\State Compute peer ensemble distillation loss $L_{pe}^{[p]}$ (Eq.\eqref{pe}) . 
		\State Compute decoupled knowledge loss $L_{dk}^{[p]}$ (Eq.\eqref{dk}). 
		\State Compute decaying ensemble loss $L_{ek}^{[p]}$ (Eq.\eqref{ek}). 
		\State Optimize $s$ with Eq.\eqref{loss}. 
		\State Optimize $t$ using EMA with Eq.\eqref{ema}
	\EndFor
\end{algorithmic}
\end{algorithm}
\section{Experiments}
This section discusses the datasets used in the experiments, implementation details, experimental settings, and compares the proposed method's performance with SOTAs to evaluate its superiority. We also perform ablation analysis of the proposed method to discuss its effectiveness and robustness.

\subsection{Datasets}
For evaluating our proposed method, experiments are conducted on several datasets, including CIFAR-10, CIFAR-100 \cite{krizhevsky2009learning} and TinyImageNet \cite{deng2009imagenet}. \textbf{CIFAR-10} consists of 50,000 training images and 10,000 validation images from 10 object categories, where each image is a $32\times 32$ RGB image. \textbf{CIFAR-100} has the same amount and the same scale of images as CIFAR-10, except that CIFAR-100 is collected from 100 object categories. \textbf{TinyImageNet} is composed of 100,000 $64\times 64$ colored images from 200 classes, with 500 training images and 50 validation images for each class. In the designed experiments, validation accuracy is reported.

\subsection{Implementation details}
All experiments are implemented based on pytorch. Models are provided by Chen et al. \cite{chen2020online}, including ResNet \cite{he2016deep}, VGG \cite{simonyan2014very}, DenseNet \cite{huang2017densely} and WideResNet \cite{zagoruyko2016wide}, and three-peer architecture is used like PCL \cite{wu2021peer}. For data augmentations, random horizontal flip, random cropping and normalization are applied in training, and images are resized to $32\times 32$ for CIFAR-10/100 and $64\times 64$ for TinyImageNet, respectively. SGD with Nesterov momentum is used to optimize the networks, where momentum is 0.9 and weight decay is $5e-4$. Batch size is set to 128, and $\gamma$ is set to 0.5. Decoupled knowledge needs one-iteration optimization with learning rate 0.01 while networks are trained for 300 epochs on CIFAR-10/100 and 100 epochs on TinyImageNet. Initial learning rate is set to 0.1, which decays by 0.1 at $\{150, 225\}$ epochs on CIFAR-10/100 and at $\{30, 60\}$ epochs on TinyImageNet.

\subsection{Comparison with SOTAs}
In this section, the proposed method DKEL is evaluated on CIFAR-10/100 and TinyImageNet datasets and compared with the previous online knowledge distillation works, including DML \cite{zhang2018deep}, CL \cite{song2018collaborative}, ONE \cite{Xu2018}, OKDDip \cite{chen2020online}, KDCL \cite{guo2020online} and PCL \cite{wu2021peer}. For fair comparisons, following \cite{Xu2018, wu2021peer}, three-branch architechures are applied in compared methods, including ONE, CL, OKDDip, PCL and DKEL, and three parallel networks are applied for the demands of DML and KDCL. For PCL, we have reproduced based on the paper since there is no open source code, and report the results of our reproduction.
\subsubsection{Results on CIFAR-10/100}
\begin{table*}[h]
	\centering
	\caption{Accuracy (\%) on CIFAR-10/100. Bold indicates the best performance. Average over 3 runs.}
	\label{tab:cifar-result}
	\resizebox{\linewidth}{!}{%
		\begin{tabular}{@{}c|c|cccccc|cc@{}}
			\toprule
			Dataset                    & Network        & DML        & CL         & ONE        & OKDDip     & KDCL       & PCL        & Baseline   & DKEL       \\ \midrule
			\multirow{5}{*}{CIFAR-10}  & ResNet-32      & 93.94±0.07 & 94.02±0.28 & 94.20±0.12 & 94.17±0.15 & 94.01±0.08 & 94.33±0.06 & 93.26±0.15 & \textbf{94.56±0.06} \\
			& ResNet-110     & 94.53±0.25 & 95.19±0.11 & 95.16±0.30 & 95.14±0.10 & 95.11±0.16 & 95.47±0.05 & 94.69±0.10 & \textbf{95.63±0.15} \\
			& VGG-16         & 94.13±0.07 & 94.14±0.15 & 94.14±0.23 & 93.98±0.06 & 94.09±0.12 & 94.28±0.44 & 93.96±0.13 & \textbf{94.89±0.10} \\
			& DenseNet-40-12 & 93.59±0.26 & 93.05±0.25 & 93.08±0.21 & 92.64±0.22 & 93.87±0.08 & 93.71±0.25 & 93.19±0.02 & \textbf{94.23±0.08} \\
			& WRN-20-8       & 95.20±0.13 & 94.59±0.08 & 94.70±0.14 & 94.83±0.15 & 95.27±0.16 & 95.42±0.05 & 94.68±0.01 &\textbf{95.53±0.06} \\ \midrule
			\multirow{5}{*}{CIFAR-100} & ResNet-32      & 73.68±0.14 & 72.33±0.46 & 73.79±0.41 & 73.25±0.38 & 73.76±0.34 & 74.14±0.05 & 71.28±0.19 & \textbf{74.83±0.12} \\
			& ResNet-110     & 77.86±0.50 & 78.83±0.58 & 78.40±0.36 & 78.54±0.26 & 78.28±0.32 & 79.59±0.38 & 76.21±0.57 & \textbf{80.01±0.32} \\
			& VGG-16         & 75.52±0.10 & 74.33±0.08 & 74.37±0.39 & 74.68±0.05 & 75.67±0.22 & 76.96±0.03 & 74.32±0.19 & \textbf{77.20±0.12} \\
			& DenseNet-40-12 & 72.60±0.31 & 71.45±0.34 & 71.60±0.38 & 71.23±0.14 & 72.52±0.42 & 72.38±0.53 & 71.03±0.15 & \textbf{72.78±0.18}          \\
			& WRN-20-8       & 79.77±0.07 & 79.40±0.12 & 79.10±0.39 & 78.83±0.06 & 79.37±0.30 & 80.71±0.22 & 78.03±0.40 & \textbf{80.80±0.27} \\ \bottomrule
		\end{tabular}%
	}
\end{table*}
As shown in Table \ref{tab:cifar-result}, DKEL improves by approximately 0.3\% in average on both CIFAR-10 and CIFAR-100 datasets with various networks compared to PCL, which proves the effectiveness of the proposed method. For example, on CIFAR-10, DKEL improves PCL by 0.61\% and 0.52\% with VGG-16 and DenseNet-40-12 respectively, and obtains 0.18\% and 0.16\% improvement with ResNet-32 and ResNet-100 respectively. On CIFAR-100, DKEL improves PCL by 0.42\% and 0.4\% with DenseNet-40-12 and ResNet-110 respectively, and obtains 0.37\% and 0.24\% improvement with ResNet-32 and VGG-16 respectively. On both CIFAR-10 and CIFAR-100, DKEL has only a small performance gain with WRN-20-8 relative to PCL, probably because it is difficult for the network near the performance limit to achieve greater gains from the ensembled decoupled knowledge.

\subsubsection{Results on TinyImageNet}
Table \ref{tab:tiny_result} shows the accuracy performance on TinyImageNet. 
DKEL improves baseline by about 13\% for both ResNet-18 and ResNet-34. Compared with PCL, DKEL obtains 0.1\% and 0.04\% improvement with ResNet-18 and ResNet-34, respectively. The reason for the insignificant DKEL improvement for ResNet-34 may be that both the dataset and the network are more complex and the number of iterations to construct decoupled knowledge is too small to provide significantly effective supervision in the early stage of training.
\begin{table}[h]
	\centering
	\caption{Accuracy (\%) on TinyImageNet. Bold indicates the best performance. Average over 3 runs.}
	\label{tab:tiny_result}
	\resizebox{0.8\linewidth}{!}{%
		\begin{tabular}{ccccccc}
			\toprule
			Method   & ResNet-18 & ResNet-34 \\ \midrule
			Baseline & 51.43±0.03     & 51.34±0.01     \\
			KD       & 53.25±0.07     & 53.70±0.06     \\
			DML      & 53.84±0.04     & 54.02±0.13     \\
			OKDDip   & 52.61±0.07     & 56.82±0.03     \\
			KDCL     & 53.91±0.04     & 53.36±0.02     \\
			PCL      & 64.23±0.08     & 64.62±0.09     \\
			DKEL    & \textbf{64.33±0.07}     & \textbf{64.66±0.09}          \\ \bottomrule
		\end{tabular}%
	}
\end{table}

\subsubsection{Ablation study}
Ablation studies on the proposed two loss functions are performed on CIFAR-10/100 using ResNet-32. Table \ref{tab:ablation} reports the results and shows the effectiveness of the proposed method. For example, decoupled knowledge improves baseline by 0.53\% on CIFAR-100 with the help of $L_{dk}$, while the decaying ensemble strategy improves baseline by 0.5\%, and both of the two designs improve the baseline by 0.69\%.
\begin{table}[h]
	\centering
	\caption{Ablation study of the proposed method on CIFAR-10/100 using ResNet-32.}
	\label{tab:ablation}
	\resizebox{\columnwidth}{!}{%
		\begin{tabular}{@{}lccc@{}}
			\toprule
			Settings             & CIFAR-10 & CIFAR-100 \\ \midrule
			Baseline             & 94.33    & 74.14     \\
			Baseline+$L_{dk}$ & 94.54    & 74.67     \\
			Baseline+$L_{ek}$          & 94.35    & 74.64     \\
			Baseline+$L_{dk}$+$L_{ek}$  & 94.56    & 74.83     \\ \bottomrule
		\end{tabular}%
	}
\end{table}

\begin{figure*}[h]
	\centering
	\includegraphics[width=0.9\linewidth]{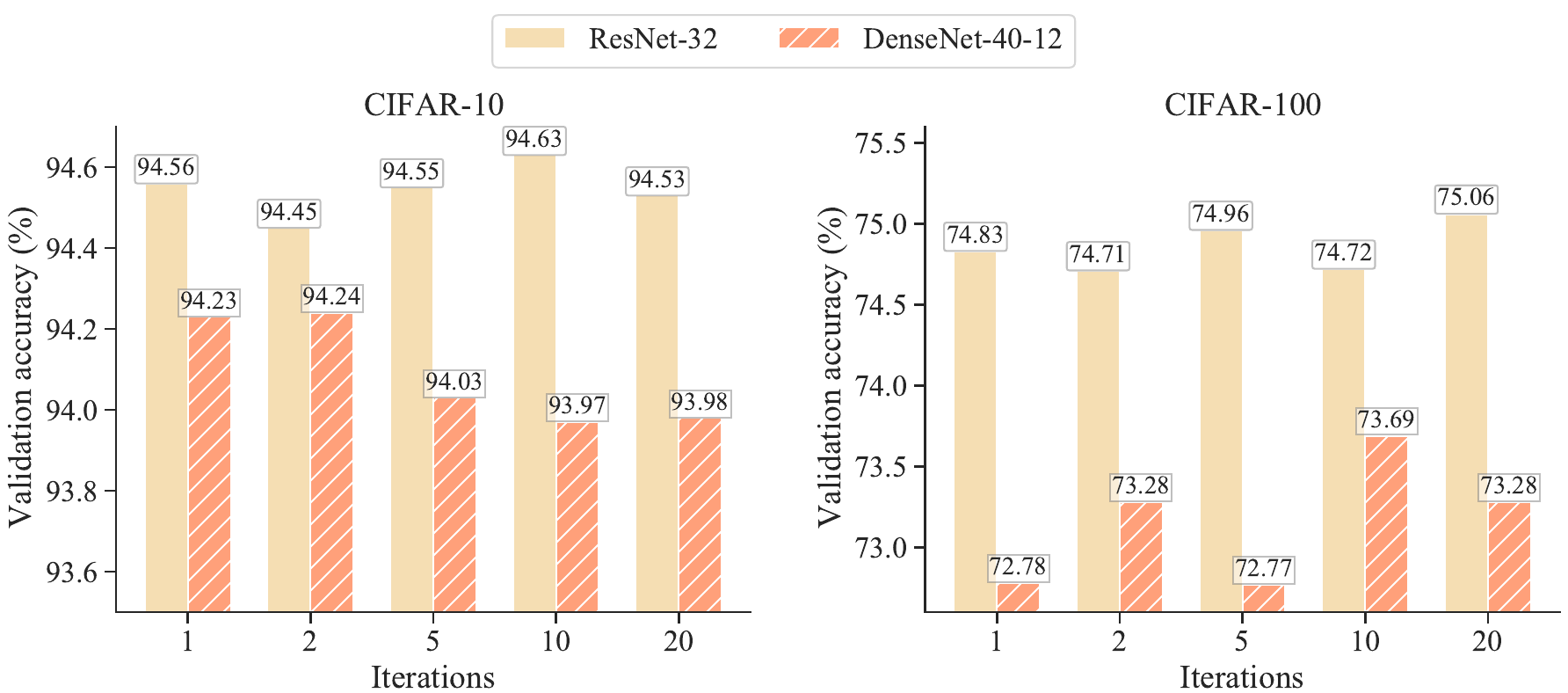}
	\caption{Illustration of network performance with different iterations for decoupled knowledge.}
	\label{fig:ablation_init}	
\end{figure*}

\textbf{Different iterations for decoupled knowledge} has an affect on the performance of the network. Fig. \ref{fig:ablation_init} illustrates a relationship between the performance of ResNet32 and DenseNet-40-12 on CIFAR-10/100 and the iterations. The left column and the right column reports the results of CIFAR-10 and CIFAR-100, respectively, and the wheat bar indicates the ResNet-32 and the salmon bar indicates the DenseNet-40-12. Three of the four settings present a positive trend, i.e. higher performance is obtained with more iterations for decoupled knowledge. The above results are basically following our expectations: 1) When the more the initial decoupled knowledge can represent the true distribution of the dataset, the more effective the knowledge learned by the original network is supposed to be. 2) Good distillation performance can be achieved with few iterations, which greatly reduces the training overhead. Specifically, 1-iteration and 10-iteration ResNet-32 obtains 94.56\% and 94.63\% performance on CIFAR-10, respectively, and 1-iteration and 5-iteration ResNet-32 achieves 74.83\% and 74.96\% accuracy on CIFAR-100, respectively.  For DenseNet-40-12 on CIFAR-10, the number of model parameters is smaller than that on CIFAR-100 as well as ResNet-32 on CIFAR-10, so it is more likely to lead to a local optimum in constructing the decoupled knowledge network, which can be improved after more iterations.

\textbf{Sensitivity analysis study on hyperparameters $\gamma$.} VGG-16 and DenseNet-40-12 are trained on CIFAR-10/100 with different $\gamma$. As shown in Table \ref{tab:exp}, VGG-16 has a performance range of 94.25 to 94.89 on CIFAR-10 and 76.73 to 77.20 on CIFAR-100, while DenseNet-40-12 has a performance range of 93.64 to 94.23 on CIFAR-10 and 72.75 to 73.36 on CIFAR-100, which means there is an performance range of about 0.6\%. Besides, when $\gamma=1.0$, networks are more likely to get poor results because $\gamma$ is too large causing the weight of $L_{ek}$ to decay too fast and converge to 0 after only two epochs of optimization. When $\gamma$ is small, the weight of $L_{ek}$ converges to 0 after dozens of training epochs and does not negatively affect the optimization objective during the later stage of training, and the Table \ref{tab:exp} also shows that smaller $\gamma$ values are beneficial for network performance.
\begin{table}[h]
	\centering
	\caption{Accuracy performance under different hyperparameter $\gamma$.}
	\label{tab:exp}
	\resizebox{\columnwidth}{!}{%
		\begin{tabular}{@{}ccccc@{}}
			\toprule
			Network & \multicolumn{2}{c}{VGG-16} & \multicolumn{2}{c}{DenseNet-40-12} \\ \midrule
			dataset & CIFAR-10    & CIFAR-100    & CIFAR-10        & CIFAR-100        \\ \midrule
			$\gamma=0.05$    & 94.62       & 76.87        & 93.95           & 73.36            \\
			$\gamma=0.10$     & 94.25       & 76.73        & 93.97           & 73.21            \\
			$\gamma=0.20$     & 94.59       & 76.74        & 93.85           & 72.82            \\
			$\gamma=0.50$     & 94.89       & 77.20        & 94.23           & 72.78            \\
			$\gamma=1.00$     & 94.79       & 76.76        & 93.64           & 72.75            \\ \bottomrule
		\end{tabular}%
	}
\end{table}

\begin{figure}[h]
	\centering
	\includegraphics[width=\linewidth]{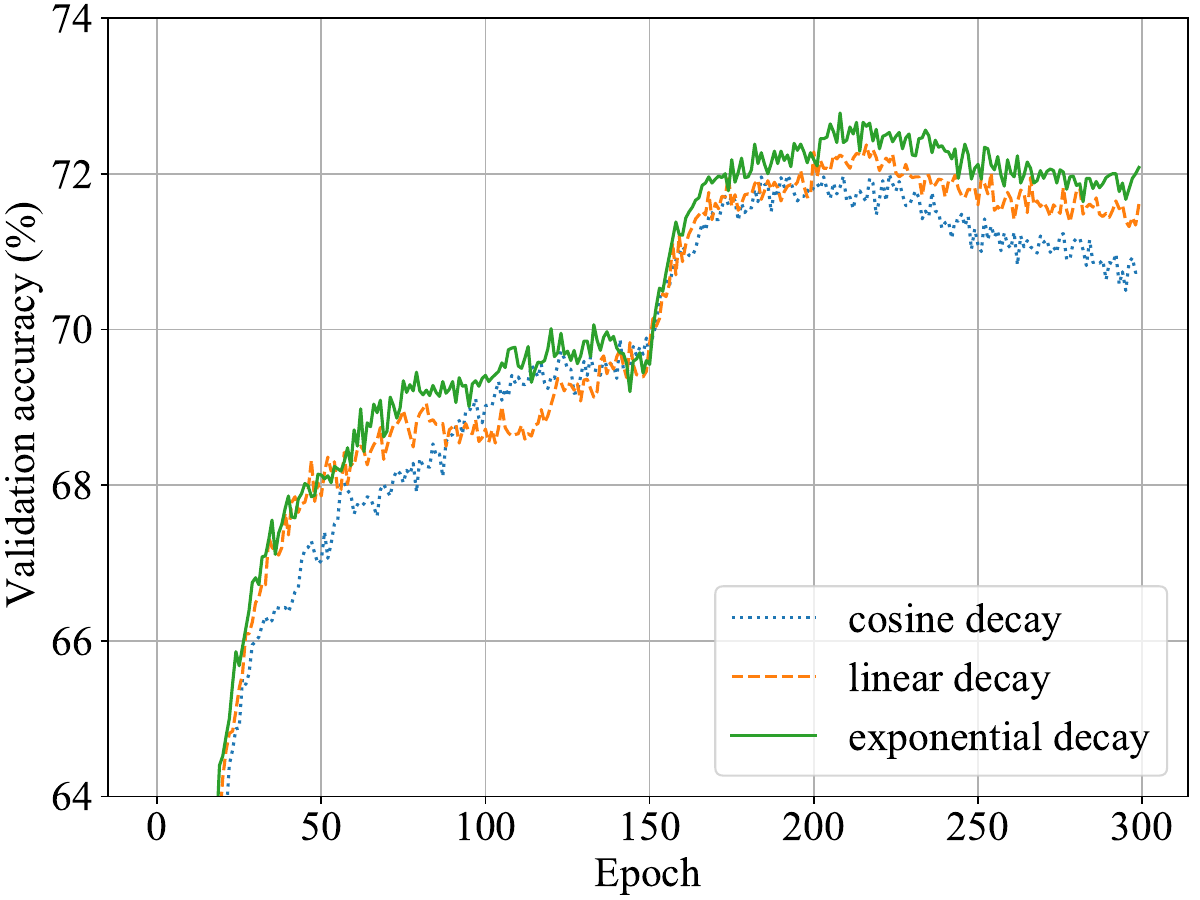}
	\caption{Validation accuracy curves with DenseNet-40-12 on CIFAR-100 using different decay schemes. The green curve indicates the exponential decay, the golden dot curve indicates the linear decay and the blue dot curve indicates the cosine decay.}
	\label{fig:decays}	
\end{figure}

\textbf{The impact of several decay schemes} for ensemble knowledge on training performance is explored. The weight of ensemble knowledge should decrease with the increase of training epoch, therefore a cosine decay and a linear decay schemes are conducted to compare with the designed exponential decay. $Epoch_{max}$ and $e$ denotes the max training epoch and the current epoch, respectively, thus the cosine decay scheme denotes:
\begin{equation}
	\omega_{cos}(e) = 0.5\times cos(\frac{\pi}{Epoch_{max}}e) + 0.5,
\end{equation}
and the linear decay scheme denotes:
\begin{equation}
	\omega_{linear}(e) = -\frac{1}{Epoch_{max}}e+1.
\end{equation}
Fig. \ref{fig:decays} shows that the designed exponential decay scheme obtains the best performance at almost all epochs while the cosine gets the worst. In terms of training stability, the exponential decay scheme is also more stable than the other two , and it can be noticed that the cosine decay scheme encounters severe overfitting leading to a significant deterioration of the performance at a later stage.

\section{Conclusion}\label{sec13}
We propose DKEL, a method for online knowledge distillation that uses decoupled knowledge generated by a temporal mean teacher to avoid model collapse. We also design a scheme to produce early decoupled knowledge and a decaying ensemble strategy to enhance early supervision robustness and reduce target deviation of late knowledge. Ideal analysis and Monte Carlo-based simulations demonstrate the motivation and mechanism, while experiments confirm the method's effectiveness and superiority.

\backmatter

\section*{Declarations}

\begin{itemize}
	\item Funding: This work was supported by the National Natural Science Foundation of China under Grant 62173160.
	\item Conflict of interest/Competing interests: All authors disclosed no relevant relationships.
	\item Availability of data and materials: The data that support the findings of this study are available from the CIFAR-10/100 and TinyImageNet datasets. CIFAR-10/100 are available in \url{https://www.cs.toronto.edu/∼kriz/cifar.html}, reference number \cite{krizhevsky2009learning}. TinyImageNet is available in \url{https://image-net.org/}, reference number \cite{deng2009imagenet}.
	\item Authors' contributions: Baitan Shao: Methodology, Software, Validation, Formal analysis, Investigation, Data Curation, Writing - Original Draft, Visualization; \\Ying Chen: Resources, Writing - Review and Editing, Supervision, Project administration, Funding acquisition
	
\end{itemize}
\bigskip





\bibliography{sn-bibliography}


\end{document}